\documentclass[runningheads]{llncs}
\usepackage[T1]{fontenc}
\usepackage{algorithm}
\usepackage{algorithmic}

\usepackage{comment}
\usepackage{amsmath,amssymb} 
\usepackage{color}
\usepackage{amsmath}
\usepackage{amssymb}
\usepackage{mathtools}
\usepackage{wrapfig}

\usepackage{graphicx}
\usepackage{amsmath,amsfonts,bm}
\usepackage{booktabs}
\usepackage{algorithm}
\usepackage{algorithmic}

\usepackage{bbding}
\usepackage{multirow}
\usepackage{subfigure}
\usepackage{booktabs}
\usepackage{array}
\usepackage[flushleft]{threeparttable}
\def\bs{\bm}

\begin{document}
\title{M-EBM: Towards Understanding the Manifolds of Energy-Based Models}
\titlerunning{M-EBM}
%
\author{Xiulong Yang\orcidID{0000-0003-3417-7106} \and
Shihao Ji}
\institute{Georgia State University \\
\email{xyang22@gsu.edu}, \email{sji@gsu.edu}}
%
%
%
\maketitle              
\begin{abstract}

    Energy-based models (EBMs) exhibit a variety of desirable properties in predictive tasks, such as generality, simplicity and compositionality. However, training EBMs on high-dimensional datasets remains unstable and expensive. In this paper, we present a Manifold EBM (M-EBM) to boost the overall performance of unconditional EBM and Joint Energy-based Model (JEM). Despite its simplicity, M-EBM significantly improves unconditional EBMs in training stability and speed on a host of benchmark datasets, such as CIFAR10, CIFAR100, CelebA-HQ, and ImageNet 32x32. Once class labels are available, label-incorporated M-EBM (M-JEM) further surpasses M-EBM in image generation quality with an over 40\% FID improvement, while enjoying improved accuracy. The code can be found in \url{https://github.com/sndnyang/mebm}.
\end{abstract}

\section{Introduction}

Energy-Based Models (EBMs) are an class of probabilistic models, which are widely applicable in image generation, out of distribution detection, adversarial robustness, and hybrid discriminative-generative modeling~\cite{nijkamp2019learning,du2019implicit,improvedCD,diffusionRecovery,jem,jempp,nomcmc,vaebm}. However, training EBMs on high-dimensional datasets remains very challenging. Most of the works utilize the Markov Chain Monte Carlo (MCMC) sampling~\cite{welling2011bayesian} to generate samples from the model distribution represented by an EBM. Specifically, they require $K$-step Langevin Dynamics sampling~\cite{welling2011bayesian} to generate samples from the model distribution in every iteration, which can be extremely expensive when using a large number of sampling steps, or highly unstable with a small number of steps. The trade-off between the training time and stability prevents the MCMC sampling based EBMs from scaling to large-scale datasets. 

\begin{figure}[ht]
    \centering
        \includegraphics[width=1.0\columnwidth]{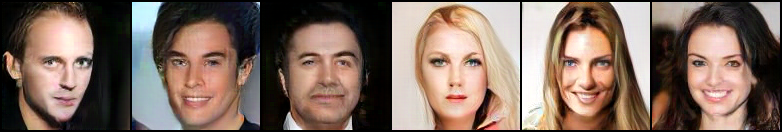}
    \vspace{-20pt}
    \caption{Generated samples of CelebA-HQ 128x128 from our M-EBM.}
    \label{figure:CelebA-HQ128_examples}
    \vspace{-20pt}
\end{figure}

Recently, there are a flurry of works on improving EBMs. The most recent studies ~\cite{improvedCD,diffusionRecovery} on the MCMC-based approach focus on improving the generation quality and stability. However, they still resort to a long sampling chain and requires expensive training. Another branch of works~\cite{nomcmc,vaebm} augment the EBM with a regularized generator in a GAN-style training to improve the stability and speed, sacrificing the desired property of learning a single object. Moreover, JEM~\cite{jem} proposes an elegant framework to reinterpret the modern CNN classifier as an EBM and achieves impressive performances in image classification and generation simultaneously. However, it also suffers from the divergence issue of the MCMC-based sampling, and its generative performance falls behind state-of-the-art EBMs. Tackling the limitations of JEM, JEM++~\cite{jempp} introduces a variety of training procedures and architecture features to improve JEM in terms of accuracy, speed and stability altogether. Furthermore, JEM++ demonstrates a trade-off between classification accuracy and image quality, but it still cannot improve image generation quality notably.

In this paper, we introduce simple yet effective training techniques to improve unconditional EBM and JEM in terms of image generation quality, training stability and speed altogether. First, the informative initialization introduced in JEM++ dramatically improves the training stability and reduces the required MCMC sampling steps. However, it's not scalable for high-resolution and large-scale datasets. Hence, we introduce a simplified informative initialization that is suitable for unconditional EBM and JEM for high-resolution images and a large number of classes (e.g., 128x128 CelebA-HQ and 1000-class ImageNet 32x32 datasets). We name our models as Manifold EBM (M-EBM) and Manifold JEM (M-JEM) respectively.  Second, we find the $L_2$ regularization of the energy magnitude does not work with the energy function utilized in JEM. To enable $L_2$ regularization and improve the training stability, we augment the standard softmax classifier with a new energy head, which is then $L_2$ regularized. Despite the simplicity, these techniques allow us to reduce the number of MCMC sampling steps of EBM dramatically, while retaining or sometimes improving classification accuracy of prior state-of-the-art EBMs. 

Our main contributions are summarised as follows:
\begin{enumerate}

\item We simplify the informative initialization in JEM++ for the SGLD chain, which stabilizes and accelerates the training of unconditional EBM and JEM, while being scalable for high-resolution and large-scale datasets. 
\item Adding an $L_2$-regularized energy head on top of a CNN feature extractor to represent an energy function stabilizes the training of JEM. Then we train M-JEM using two mini-batches: one with data augmentation for classification, and the other one without data augmentation for maximum likelihood estimation of EBMs.
\item We conduct extensive experiments on four benchmark datasets. M-EBM matches or outperforms prior state-of-the-art unconditional EBMs, while significantly improves training stability and reduces the number of sampling steps. Moreover, M-JEM improves JEM's training stability and speed, image generation quality, and classification accuracy altogether, while outperforming M-EBM in image generation quality. 
\end{enumerate}

\section{Background}

Energy-based Models (EBMs)~\cite{lecun2006tutorial} utilizes the idea that any probability density $p_{\bs{\theta}}(\bs{x})$ can be expressed as
\begin{equation}\label{eq:ebm_define}
  p_{\bs{\theta}}(\bs{x})=\frac{\exp \left(-E_{\bs{\theta}}(\bs{x})\right)}{Z(\bs{\theta})},
\end{equation}
where $E_{\bs{\theta}}(\bs{x})$ is named the energy function that maps each input $\bs{x}\in\mathcal{X}$ to a scalar, and $Z(\bs{\theta}) = \int_{\bs{x}} \exp \left(-E_{\theta}(\bs{x})\right)d\bs{x}$ is the normalizing constant w.r.t $\bs{x}$ (also known as the partition function). Ideally, an energy function should assign low energy values to samples drawn from data distribution and high values otherwise.

The key challenge of EBM training is estimating the intractable partition function $Z(\bs{\theta})$, and the maximum likelihood estimation of parameters $\bs{\theta}$ is not straightforward. A number of sampling-based approaches have been proposed to approximate the partition function effectively. Specifically, the derivative of the log-likelihood of $\bs{x}\in\mathcal{X}$ w.r.t. $\bs{\theta}$ can be expressed as
\begin{align}\label{eq:ml}
  \frac{\partial\log p_{\bs{\theta}}(\bs{x})}{\partial\bs{\theta}}   =   \mathbb{E}_{p_{\bs{\theta}}(\bs{x}')}  \left[ \frac{\partial E_{\bs{\theta}}(\bs{x}')}{\partial\bs{\theta}} \right]
  -  \mathbb{E}_{p_d(\bs{x})}  \left[ \frac{\partial E_{\bs{\theta}}(\bs{x})}{\partial\bs{\theta}} \right],
\end{align}
where the first expectation is over the model density $p_{\bs{\theta}}(\bs{x}')$, which is challenging due to the intractable $Z(\bs{\theta})$. 

To estimate it efficiently, MCMC and Gibbs sampling~\cite{hinton2002cd} have been proposed. Moreover, to speed up the sampling, recently Stochastic Gradient Langevin Dynamics (SGLD)~\cite{welling2011bayesian} is employed to train EBMs~\cite{nijkamp2019learning,du2019implicit,jem}.  Specifically, to sample from $p_{\bs{\theta}}(\bs{x})$, the SGLD follows
\begin{align}\label{eq:sgld}
    &\bs{x}^0\sim p_0(\bs{x}), 
    &\bs{x}^{t+1} = \bs{x}^t-\frac{\alpha}{2} \frac{\partial
    E_{\bs{\theta}}(\bs{x}^t)}{\partial \bs{x}^t} + \alpha\epsilon^t, \;\; \epsilon^t \sim \mathcal{N} (0,1),
\end{align}
where $p_0(\bs{x})$ is typically a uniform distribution over $[-1,1]$, whose samples are refined via a noisy gradient decent with step-size $\alpha$ over a sampling chain. 

Prior works~\cite{nijkamp2019anatomy,nijkamp2019learning,du2019implicit,jem} have investigated the effect of hyper-parameters in SGLD sampling and showed that the SGLD-based approaches suffer from poor stability and prolonged computation of sampling at every iteration. Nijkamp et al.~\cite{nijkamp2019anatomy} find that it's desirable to generate samples from the SGLD chain after it converges. The convergence requires the step-size $\alpha$ to decay with a polynomial schedule and infinite sampling steps, which is impractical. Therefore, Short-Run and Long-Run MCMC samplings are utilized for EBM training. Moreover, most works~\cite{du2019implicit,jem,improvedCD} use a constant step-size $\alpha$ during sampling and approximate the samples with a sampler that runs only for a finite number of steps, which is still computationally very expensive. Another recent work~\cite{diffusionRecovery} combines the SGLD-based approach with diffusion models~\cite{DDPM} under a framework of conditional EBMs. They achieve state-of-the-art image generation quality and obtain a faithful energy potential.

Joint Energy-based Models (JEM)~\cite{jem} demonstrates that standard softmax-based classifiers can be trained as EBMs. Given an input $\bs{x} \in R^{D}$, a classifier of parameters $\bs{\theta}$ maps the input to a vector of $C$ real-valued numbers (known as logits): $f_{\bs{\theta}}(\bs{x})[y], \forall y\in [1,\cdots,C]$, where $C$ is the number of classes. Then the softmax function is employed to convert the logits into a categorical distribution: $p_{\bs{\theta}}(y|\bs{x})=e^{f_{\bs{\theta}}(\bs{x})[y]} / \sum_{y^{\prime}} e^{f_{\bs{\theta}}(\bs{x})\left[y^{\prime}\right]}$. The authors reuse the logits to define an energy function for the joint density: $p_{\bs{\theta}}(\bs{x}, y)=e^{f_{\bs{\theta}}(\bs{x})[y]} / Z(\bs{\theta})$. Then a marginal density of $\bs{x}$ can be achieved by marginalizing out $y$ as: $p_{\bs{\theta}}(\bs{x})=\sum_{y} p_{\bs{\theta}}(\bs{x}, y) = \sum_{y} e^{f_{\bs{\theta}}(\bs{x})\left[y\right]} / Z(\bs{\theta})$.
As a result, the corresponding energy function of $\bs{x}$ is defined as 
\begin{equation}\label{eq:jem_ex}
    E_{\bs{\theta}} (\bs{x})\!=\!-\log\! \sum_{y}\!e^{f_{\bs{\theta}}(\bs{x})\left[y\right]}\!=\!-\text{LSE}( f_{\bs{\theta}}(\bs{x})),
\end{equation}
where $\text{LSE}(\cdot)$ denotes the Log-Sum-Exp function. The advantage of this LSE energy function is that an additional degree of freedom in the scale of the logit vector now can model the data distribution.

To optimize the model parameter $\bs{\theta}$, JEM maximizes the logarithm of joint density function $p_{\bs{\theta}}(\bs{x},y)$:
\begin{equation}\label{eq:jem_loss}
  \log p_{\bs{\theta}}(\bs{x}, y) = \log p_{\bs{\theta}}(y|\bs{x}) + \log p_{\bs{\theta}}(\bs{x}),
\end{equation}
where the first term is the cross-entropy objective for classification, and the second term is the maximum likelihood learning of EBM as shown in Eq.~\ref{eq:ml}. We can also interpret the second term as an unsupervised regularization on the model parameters $\bs{\theta}$.

\section{Manifold EBM}

\subsection{Informative Initialization and M-EBM}

As shown in Eq.~\ref{eq:sgld}, the SGLD sampling starts from an initial distribution $p_0(\bs{x})$. To train the EBM as a generative model, Short-Run MCMC sampling~\cite{nijkamp2019learning} utilizes an MCMC sampler that starts from a random noise distribution such as a uniform distribution. A concurrent work IGEBM~\cite{du2019implicit} proposes an initialization approach with a sample replay buffer in which they store past generated samples and draw samples from either replay buffer or uniform random noise to initialize the Langevin dynamics procedure. This is also the sampling approach adopted by~\cite{jem,jeat}. Furthermore, JEM++~\cite{jempp} introduces an informative initialization with the replay buffer by using a Gaussian mixture distribution estimated from the training images, which significantly reduces the number of sampling steps required by SGLD while improving its training stability.

However, the per-class covariance matrices of the Gaussian mixture distribution utilized by JEM++ can be huge for high-resolution image datasets with a large number of classes. Hence, we estimate a single Gaussian distribution from the whole training dataset. That is, we estimate the initial sampling distribution as
\begin{align}\label{eq:p0}
  p_0(\bs{x})&=\mathcal{N}(\bs{\mu}, \bs{\Sigma})\\
  \text{with}\quad 
\bs{\mu}  = \mathbb{E}_{\bs{x} \sim \mathcal{D} }[\bs{x}],  \quad 
\bs{\Sigma} &=\mathbb{E}_{\bs{x} \sim \mathcal{D} }\left[\left( \bs{x} - \bs{\mu} \right)\left( \bs{x} -\bs{\mu} \right)^{\top}\right],\nonumber
\end{align}
where $\mathcal{D}$ denotes the whole training set. The visualization of the estimated centers and samples from $p_0(\bs{x})$ of different datasets are provided in the appendix. Since only one Gaussian distribution is estimated from the whole training set, we can apply it for unconditional datasets such as CelebA, and reduce the memory and space required for the large covariance matrices\footnote{One covariance matrix of CIFAR10 has $(3\times 32 \times 32)^2 \approx 9.4M$ parameters and uses 37.6MB memory. A dataset with $C$ classes will take $37.6 \times C$ MB.}. Although $\bs{\mu}$ and $\bs{\Sigma}$ can be well estimated with sufficient samples, they still lead to a biased initialization with higher variance, compared to the Gaussian mixture initialization utilized in JEM++. But our empirical study shows that our simplified initialization won't deteriorate the performance and is comparable to bias-reduced Gaussian mixture initialization.

Since the manifold of $\bs{x}_0$ from our informative initialization is much closer to the real data manifold than that of uniform initialization, this informative initialization reduces the required sampling steps (and thus accelerates training), and also improves training stability as we will demonstrate in the experiments. We therefore call the EBM with this simplified informative initialization as M-EBM throughout this work.

\subsection{Injected Noise in M-EBM}

Existing work~\cite{nijkamp2019learning} studied the effect of injected noise on training stability via smoothing $p_{data}$ with additive Gaussian noises $\bs{x} \leftarrow \bs{x} + \bs{\epsilon}, \bs{\epsilon} \sim \mathcal{N}(0, \sigma^2 I)$. Their results demonstrated that the fidelity of the examples in terms of IS and FID improves, when lowering $\bs{\sigma}^2$. And they depict the tradeoff between the sampling steps $K$ and the level of injected noise, indicating the training time and the stability. After it, several following methods~\cite{diffusionRecovery,improvedCD,febm2020} successfully remove the injected noise and achieve better image quality. However, they require a very large $K\geq 30$ to stabilize the training. Thanks to the informative initialization, it not only allows us to significantly reduce $K$, but also removes the injected noise to improve the image quality while keeping high stability.  As shown in Fig.~\ref{fig:tsne_reg}, the manifolds of real data and  $\bs{x}_0$ sampled from informative initialization are very close, even mixing together when M-EBM is trained without energy regularization. Hence, we suppose the gradients  $\nabla_{\bs{x}} E(\bs{x})$ are defined (almost) everywhere in such manifolds and thus can reduce the perturbation with noise which is originally explained in NCSN~\cite{song2019generative}.

\section{Manifold JEM}\label{sec:twobatch}

\subsection{Injected Noise in M-JEM}

As discussed in previous section, the injected noise smoothing $p_{data}$  would hurt the generative performance of EBMs. For JEM and JEM++, we suppose it would also decrease the classification accuracy. Hence, it's critical to remove the injected noise and gain benefits in terms of classification accuracy and generation quality.  We use $N$ to denote the  noise-adding operation. Then the actual objective of JEM is
\begin{equation}\label{eq:pxy_da}
   \log p_{\bs{\theta}}(N(\bs{x}), y) = \log p_{\bs{\theta}}(y|N(\bs{x})) + \log p_{\bs{\theta}}(N(\bs{x})).
\end{equation}
Interestingly, we find that if we only remove the injected noise, the training is not stable. However, if we further disable the data augmentation when learning maximum likelihood $\log p_{\bs{\theta}}(\bs{x})$, it becomes even more stable than JEM++ and enjoys improved accuracy and better sampling quality. Following the observation, we train our
M-JEM using two mini-batches: one with data augmentation for classification, and the other one without data augmentation for maximum likelihood
estimation of EBMs.

\subsection{Energy Function Regularization in M-JEM}

IGEBM~\cite{du2019implicit} finds that constraining the Lipschitz constant of the energy network can ease the instability issue in Langevin dynamics. Hence, they weakly $L_2$ regularize energy magnitudes for both positive and negative samples to the contrastive divergence as:
\begin{align}\label{eq:l2_reg}
\mathcal{L} = \frac{1}{B} \sum_{i=1}^B \left( E_i^+ - E_i^- + \alpha ( {E_i^+}^2 + {E_i^-}^2 ) \right),
\end{align}
where $E^+ = E_{\bs{\theta}}(\bs{x}^+)$ with $\bs{x}^+$ sampled from the data distribution $p_{d}$, and $E^- = E_{\bs{\theta}}(\bs{x}^-)$ with $\bs{x}^-$ sampled from the model distribution $p_{\bs{\theta}}(\bs{x})$. The effect of $L_2$ regularization on EBMs can be viewed as Fig~\ref{fig:tsne_reg}. However, since $L_2$ regularization would force the vector of logits $f_{\bs{\theta}}(\bs{x})$ to be uniform, while maximizing  $p_{\bs{\theta}}(y|\bs{x})$ boosts $f_{\bs{\theta}}(\bs{x})[y]$. Hence, the $L_2$ regularization is incompatible with Eq.~\ref{eq:jem_ex} and cannot be directly applied to vanilla JEM.

\begin{figure}[t]
    \vspace{-5px}
    
    \subfigure[]{
        \includegraphics[width=0.6\textwidth]{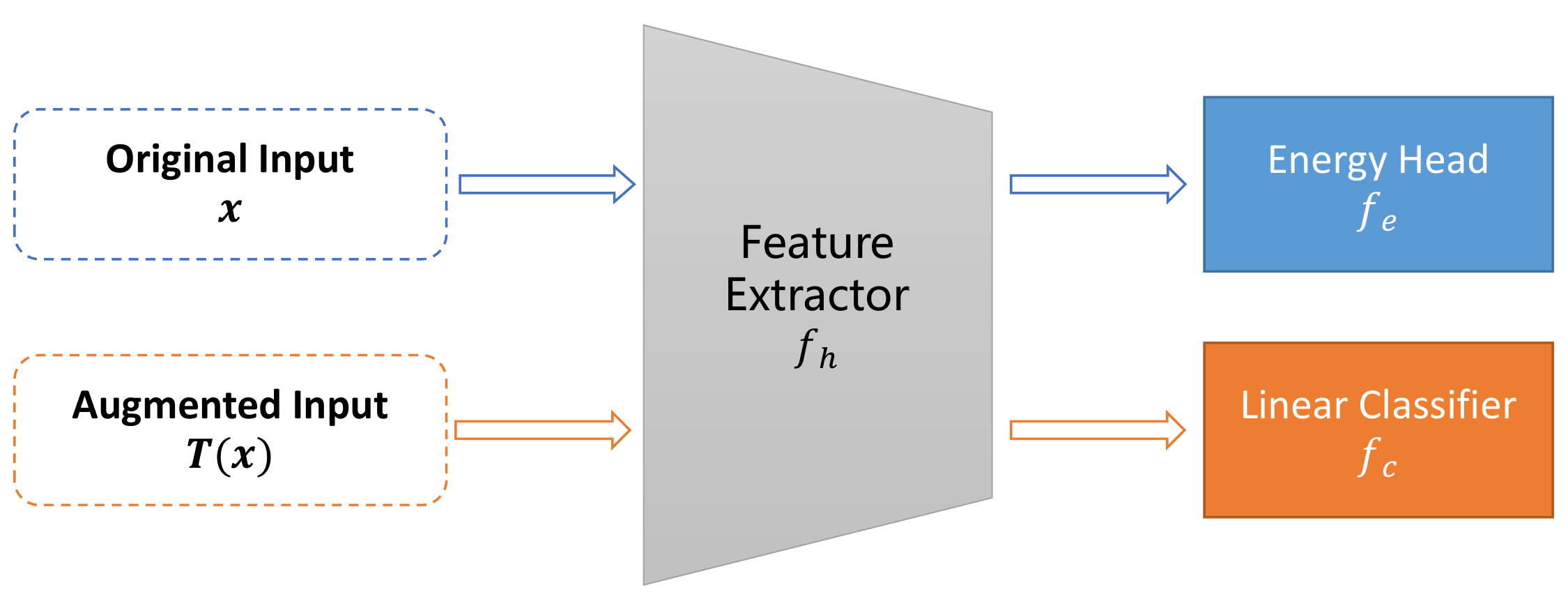}
        \label{fig:arch}
    }\vspace{-8pt}
    \subfigure[]{
        \includegraphics[width=0.36\textwidth]{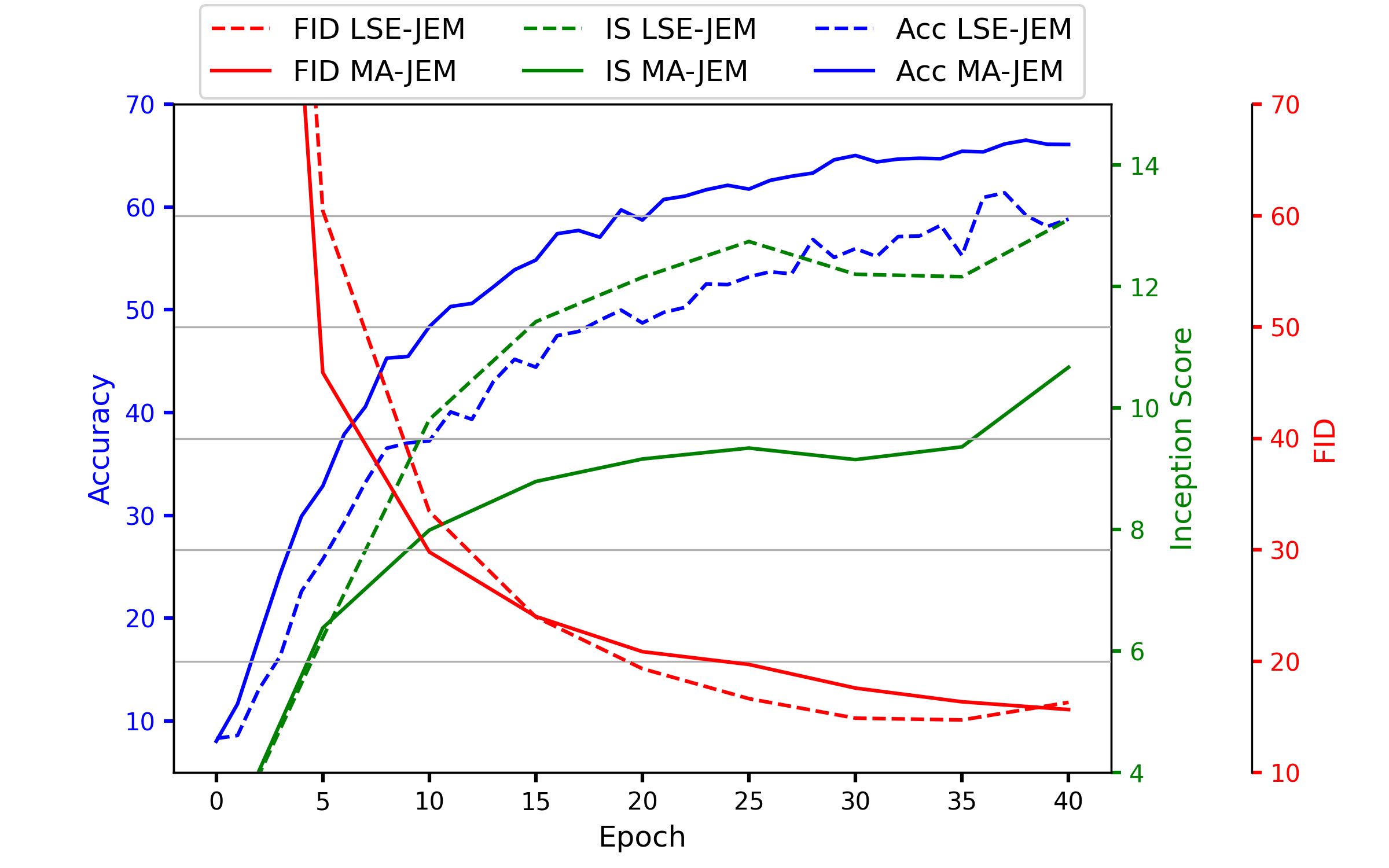}
        \label{fig:curves_cond_uncond}
    }\vspace{-8pt}
    
    \caption{a) The architecture of M-JEM. An energy head $f_e$ is augmented for energy magnitude regularization and two mini-batches are used for the training of classifier and the maximum likelihood estimate of EBM, respectively. b) Comparison between M-JEM and LSE-JEM on CIFAR100.}
    \vspace{-8px}
\end{figure}


To incorporate $L_2$ regularization to JEM, we propose to augment the standard CNN softmax classifier with an extra fully connected layer, called Energy Head, as shown in Fig.~\ref{fig:arch}. Then the $L_2$ regularization is applied on the energy head (instead of the LSE classification head) to improve the training stability.

We provide the pseudo-code for M-EBM/JEM as in Algorithm~\ref{algo:1}, which follows the framework of IGEBM~\cite{du2019implicit} and JEM~\cite{jem}.

\begin{algorithm}[ht!]
\caption{M-EBM/JEM Training: Given network $f_\theta$, SGLD step-size $\alpha$, SGLD noise $\sigma$, replay buffer $B$, SGLD steps $K$, reinitialization frequency $\rho$}
\label{algo:1}
\begin{algorithmic}[1]
\WHILE{not converged}
\STATE Sample $\bs{x}^+$ and $y$ from dataset
\STATE Sample $\widehat{\bs{x}}_0 \sim B$ with probability $1-\rho$,  else $\widehat{\bs{x}}_0 \sim p_0(\bs{x})$ as Eq.~\ref{eq:p0}  
\FOR{$t \in [1, 2, \ldots, K]$}   
  \STATE $\widehat{\bs{x}}_t = \widehat{\bs{x}}_{t-1} - \alpha \cdot \frac{\partial E(\widehat{\bs{x}}_{t-1})}{\partial \widehat{\bs{x}}_{t-1}} + \sigma \cdot \mathcal{N}(0, I)$
\ENDFOR
\STATE $\bs{x}^- = \text{StopGrad}(\widehat{\bs{x}}_K)$
\STATE $L_{\text{gen}}(\theta) = E(\bs{x}^+) - E(\bs{x}^-) + \alpha \left( E(\bs{x}^+)^2 +  E(\bs{x}^-)^2 \right)$ as Eq.~\ref{eq:l2_reg}.
\STATE $L(\theta) = L_{\text{gen}}(\theta)$ for M-EBM
\STATE $L(\theta) = L_\text{clf}(\theta) +  L_{\text{gen}}(\theta)$ with  $L_{\text{clf}}(\theta) = \text{xent}(f_\theta(\bs{x}), y)$ for M-JEM
\STATE Calculate gradient $\frac{\partial L(\theta)}{\partial \theta}$ to update $\theta$
\STATE Add $\bs{x}^-$ to $B$
\ENDWHILE
\end{algorithmic}
\end{algorithm}

\section{Experiments}

In this section, we first evaluate the generative performance of M-EBM on multiple datasets, including CIFAR10, CIFAR100, CelebA-HQ 128x128 and ImageNet 32x32. Then, we investigate the efficacy of the M-JEM on CIFAR10 and CIFAR100.  Finally, the study and visualization of the differences between trained EBM and JEM are provided to analyze their generative capability.

Our code is largely built on top of JEM~\cite{jem}\footnote{\url{https://github.com/wgrathwohl/JEM}}. 
For a fair comparison with JEM, we update each model with 390 iterations in 1 epoch. Empirically, we find a batch size of 128 for $p_{\bs{\theta}}(y|\bs{x})$ achieves the best classification accuracy on CIFAR10, while we use 64, the same batch size as in JEM, for $p_{\bs{\theta}}(\bs{x})$. We train our models on ImageNet 32x32 for 50 epochs and other datasets for 150 epochs at most. All our experiments are performed with PyTorch on Nvidia GPUs. For CIFAR10 and CIFAR100, we train the backbone Wide-ResNet 28-10~\cite{wideresnet16} on a single GPU. Due to limited computational resources, we use Wide-ResNet 28-2 for ImageNet 32x32 on a single GPU, and Wide-ResNet 28-5 for CelebA-HQ 128x128 on 2 GPUs. 

\begin{minipage}{0.5\linewidth}

\begin{threeparttable}
\caption{Inception and FID scores of M-EBM on CIFAR10.}
\label{table:unconditional_results}
\begin{tabular}{lcc}
\toprule
 Model  &  IS $\uparrow$ & FID $\downarrow$ \\
\midrule
M-EBM(K=1)*     & 6.02 & 35.7    \\
M-EBM(K=2)      & 6.72 & 27.1    \\
M-EBM(K=5)      & 7.14 & 22.7    \\
M-EBM(K=10)     & 7.08 & 20.4    \\
M-EBM(K=20)     & 7.20 & 21.1    \\
\midrule
\multicolumn{3}{l}{Explicit EBM(Unconditional)} \\
\midrule 
ShortRun(K=100)~\cite{nijkamp2019learning}   &  6.72  & 32.1 \\
IGEBM(K=60)~\cite{du2019implicit}            &  6.78  & 38.2 \\
f-EBM(K=60)~\cite{febm2020}            &  8.61  & 30.8 \\
CF-EBM(K=50)~\cite{cfebm}                    &   -    & 16.7 \\
KL-EBM(K=40)~\cite{improvedCD}               &  7.85  & 25.1 \\
DiffuRecov(K=30)~\cite{diffusionRecovery}    &  8.31  & 9.58 \\
\midrule
Regularized Generator \\
\midrule
 GEBM~\cite{generalebm}                      & -     &  23.02 \\
 VAEBM(K=6)~\cite{vaebm}                     & 8.43  &  12.19 \\
\midrule 
Other \\
\midrule
 SNGAN~\cite{miyato2018spectral}          & 8.59  & 21.7  \\
 NCSN~\cite{song2019generative}           & 8.91  & 25.3  \\ 
 StyleGAN2-ADA~\cite{styleganADA}         & \textbf{9.74}  & \textbf{2.92}  \\
 DDPM~\cite{DDPM}                         & 9.46  & 3.17  \\ 
\bottomrule
\end{tabular}
\begin{tablenotes}
  \scriptsize\item * M-EBM diverges with $K=1$, and we report the best FID before diverging. 
\end{tablenotes}
\end{threeparttable}

\end{minipage}
\begin{minipage}[ht]{0.5\linewidth}

\begin{threeparttable}
\caption{FID results of M-EBM on CIFAR100, CelebA-HQ 128, and ImageNet 32x32.}
\label{table:uncond_other_results}
\begin{tabular}{lc}
\toprule
 Model  &  FID $\downarrow$ \\
\midrule
\multicolumn{2}{l}{CIFAR100 Unconditional}  \\
\midrule
M-EBM(K=1)*      & 45.5    \\  
M-EBM(K=2)       & 26.2    \\  
M-EBM(K=5)       & 27.2    \\       
M-EBM(K=10)      & 26.9    \\   
SNGAN~\cite{miyato2018spectral}           & 22.4    \\
\midrule
\multicolumn{2}{l}{CelebA-HQ 128 Unconditional}  \\
\midrule
M-EBM(K=5)*      & 57.76    \\       
M-EBM(K=10)      & 39.87    \\ 
KL-EBM(K=40)~\cite{improvedCD}    & 28.78    \\  
SNGAN~\cite{miyato2018spectral}           &  24.36    \\
\midrule
\multicolumn{2}{l}{ImageNet 32x32 Unconditional}  \\
\midrule 
M-EBM(K=2)         &   54.52  \\       
M-EBM(K=5)         &   52.71  \\      
IGEBM(K=60)~\cite{du2019implicit}       &  62.23   \\
KL-EBM(K=40)~\cite{improvedCD}          &  32.48   \\
\bottomrule
\end{tabular}
\begin{tablenotes}
  \scriptsize\item $^*$ Our models diverge during training with given $K$, and we report the best FID before diverging. 
\end{tablenotes}
\end{threeparttable}

\end{minipage}

\subsection{M-EBM}

We first evaluate the performance of M-EBM on CIFAR10, CIFAR100, CelebA-HQ 128 and ImageNet 32x32. We utilize the Inception Score (IS)~\cite{imprgan16} and Fr\'{e}chet Inception Distance (FID)~\cite{heusel2017gans} to evaluate the quality of generated images.  

The results are reported in Table~\ref{table:unconditional_results} and~\ref{table:uncond_other_results}, respectively. It can be observed that our method consistently surpasses existing methods in terms of sampling steps by a significant margin. On CIFAR10, M-EBM outperforms many EBM approaches and SNGAN in terms of FID, while the performance is slightly worse than SNGAN on CIFAR100. Some EBM approaches show better performance, such VAEBM, CF-EBM and DiffuRecov. However, they require an extra pretrained generator, or special architecture, or much larger sampling steps, while M-EBM can train on a classical architecture as the backbone with least $K$. On ImageNet 32x32, we note that M-EBM  with $K=2$ is incredibly stable and achieves FID 54.52 within 30 epochs and outperforms IGEBM. In addition, increasing sampling steps $K$ further doesn't have an obvious improvement. 
Finally, on CelebA-HQ, M-EBM is worse than baseline methods as we find it is less stable and requires more sampling steps due to the high resolution of CelebA-HQ. Nevertheless, our method builds a new solid baseline on different large-scale benchmarks for further investigations of EBM training in these more challenging tasks. Samples generated by M-EBMs for CIFAR10, CIFAR100, and CelebA-HQ are shown in Fig.~\ref{figure:MA_JEM_samples} and Fig.~\ref{figure:CelebA-HQ128_examples}, respectively. The generated samples of ImageNet 32x32 can be found in the appendix.

\subsection{M-JEM}

\begin{figure}[hb!]

    \centering
    \subfigure[CIFAR10]{
        \includegraphics[width=0.22\columnwidth]{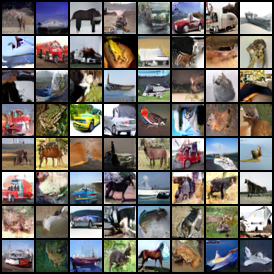}
        \label{figure:cifar10}
    }
    \subfigure[CIFAR100]{
        \includegraphics[width=0.22\columnwidth]{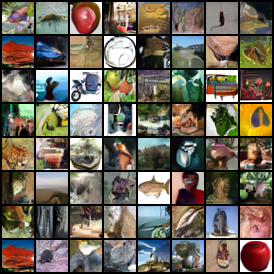}
        \label{figure:CIFAR100}
    }
    \subfigure[CIFAR10]{
        \includegraphics[width=0.22\columnwidth]{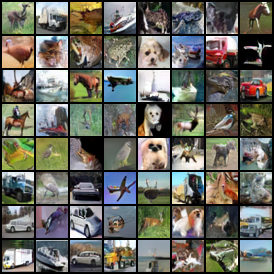}
        \label{figure:cifar10_jem}
    }
    \subfigure[CIFAR100]{
        \includegraphics[width=0.22\columnwidth]{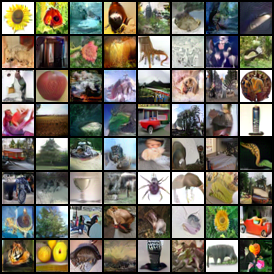}
        \label{figure:CIFAR100_jem}
    }
    \vspace{-10pt}
    \caption{M-EBM and M-JEM generated samples of CIFAR10 and CIFAR100. }
    \label{figure:MA_JEM_samples}
    \vspace{-5pt}
\end{figure}

We train M-JEM on two benchmark datasets: CIFAR10 and CIFAR100, and compare its performance to the state-of-the-art hybrid models and some representative generative models. Table~\ref{table:hybrid_results} and~\ref{table:hybrid_cifar100} report results on CIFAR10 and CIFAR100, respectively. As we can see, M-JEM improves JEM's image generation quality, stability, speed, and accuracy by a notable margin. It also boosts the IS and FID scores over M-EBM. Compared with JEM++, FID of M-JEM drops dramatically since we exclude the noise, and the notable gain of accuracy when $K=5$ indicates M-JEM($K=5$) is much more stable than JEM++($K=5$).  On CIFAR100, IS and FID scores are not commonly reported by state-of-the-art hybrid models, such as JEM~\cite{jem}, VERA~\cite{nomcmc}, and JEM++~\cite{jempp}. Hence, our work builds a baseline for hybrid modeling on CIFAR100 with decent classification accuracy and image generation quality for future investigations.  Images generated by M-JEM for CIFAR10 and CIFAR100 are can be found in Fig.~\ref{figure:MA_JEM_samples}.

\subsection{Analysis}

\subsubsection{Is Energy Head better than LSE?}

To evaluate the effect of the energy head, we conduct an experiment comparing M-JEM (with energy head) and LSE-JEM (without energy head) on CIFAR100. Fig.~\ref{fig:curves_cond_uncond} shows that M-JEM achieves much higher classification accuracy, comparable FID but a lower Inception Score than LSE-JEM. However, we empirically find LSE-JEM is less stable than M-JEM after 40 epochs  which leads us to analyze the manifolds learned by different models.

\subsubsection{Manifold Analysis}

\begin{figure}[t]

    \subfigure[]{
        \includegraphics[width=0.45\textwidth]{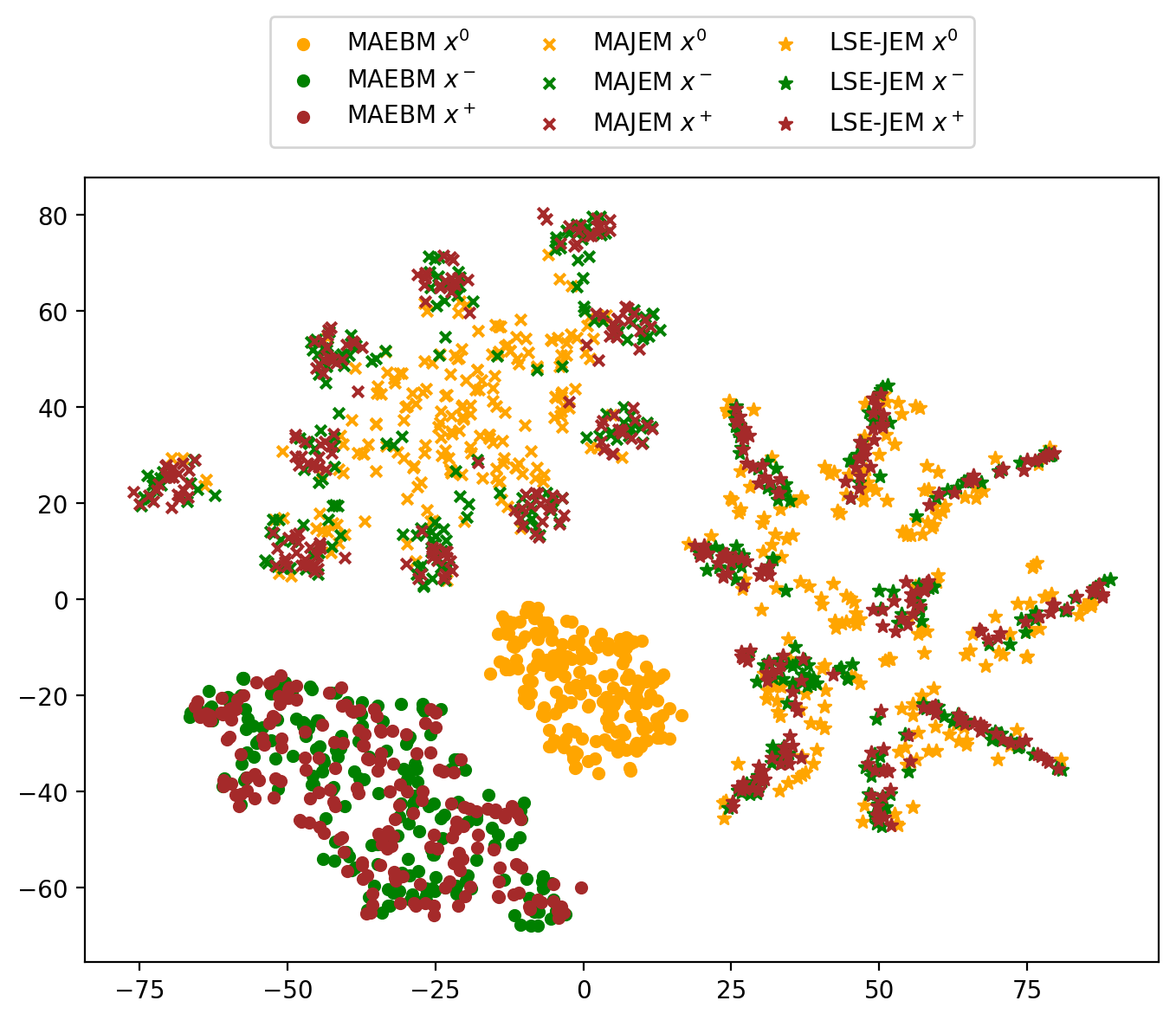}
        \label{fig:tsne}
    }
    \subfigure[]{
        \includegraphics[width=0.45\textwidth]{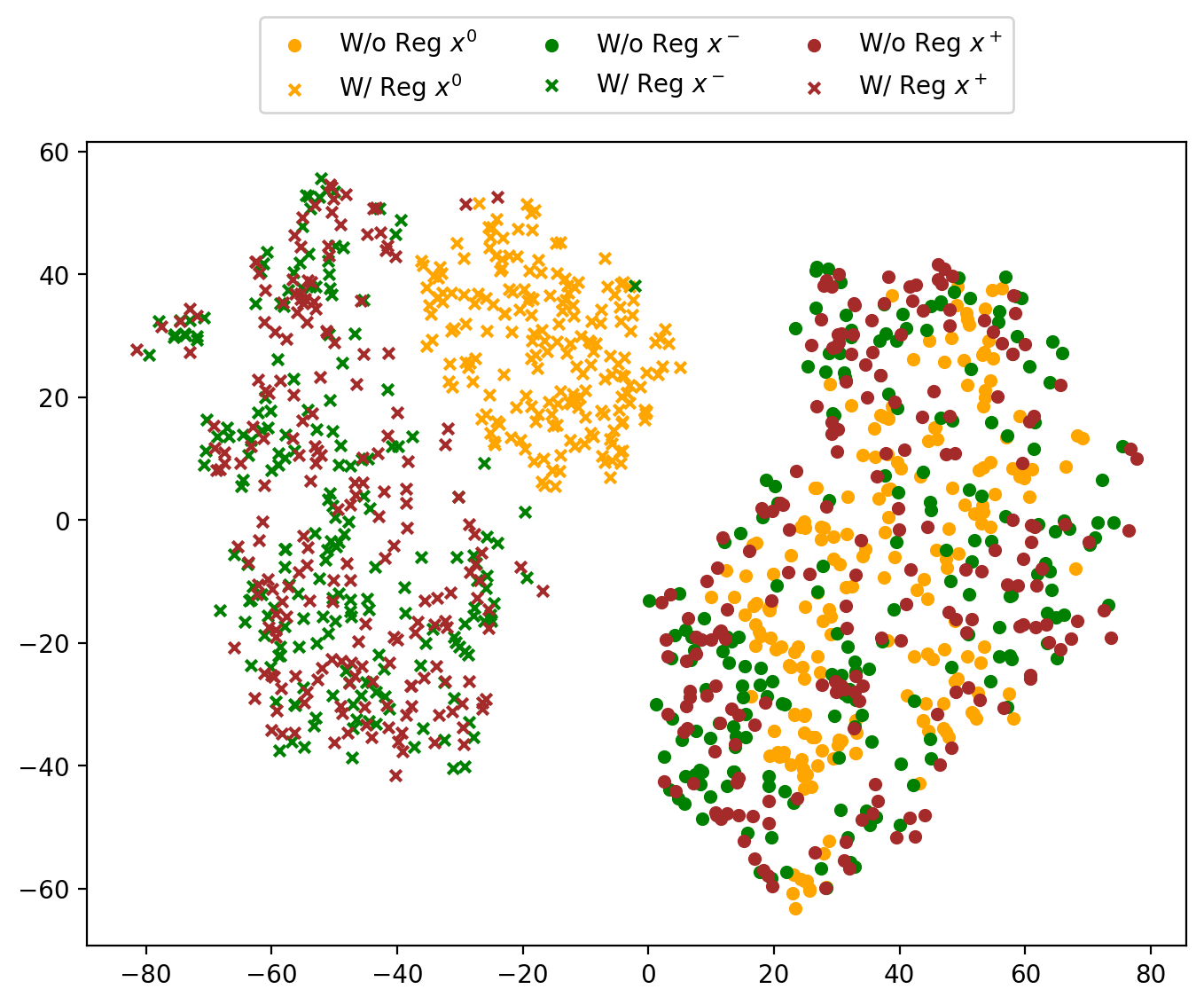}
        \label{fig:tsne_reg}
    }
    \vspace{-10pt}
    \caption{t-SNE visualization of the latent feature spaces learned by different models trained on CIFAR10. We use different colors to represent ($x^0$ initial samples, $x^-$ samples in replay buffer, $x^+$ real data), and different shapes($\circ, \times, \star$) to indicate different EBMs.}
    \label{fig:tsne_visualization}
    \vspace{-10pt}
\end{figure}

To facilitate better understanding of different approaches, we utilize the t-SNE visualization for manifold analysis shown in Fig.~\ref{fig:tsne} and ~\ref{fig:tsne_reg}. To have a fair comparison, we pick fixed samples from CIFAR10 as $\bs{x}^+$, initialize samples from $p_0(\bs{x})$ as $\bs{x}^0$, and randomly select samples from the replay buffer of each pre-trained models as $\bs{x}^-$. Given the inputs from $\bs{x}^+$, $\bs{x}^-$ and $\bs{x}^0$,  we collect the outputs of the penultimate layer as features and apply the t-SNE technique to generate the visualization. For Fig~\ref{fig:tsne}, three CIFAR10-trained M-EBM, M-JEM, and LSE-JEM with $K=10$ are involved. We further conduct the comparison between M-EBMs($K=5$) with and without energy $L_2$ regularization in Fig~\ref{fig:tsne_reg}. 



\begin{minipage}[ht]{0.47\linewidth} 

\begin{threeparttable}
\caption{Hybrid Modeling Results on CIFAR10.}
\label{table:hybrid_results}
\begin{tabular}{lccc}
\toprule
 Model  & Acc \% $\uparrow$ & IS $\uparrow$ & FID $\downarrow$ \\
 \midrule
M-JEM(K=1)*   & 78.4 & 7.91 & 29.8    \\
M-JEM(K=2)*   & 86.5 & 8.64 & 19.3    \\
M-JEM(K=5)   & 93.1 & 8.71 & 12.1    \\
M-JEM(K=10)  & 93.8 & 8.52 & \textbf{11.5} \\
M-JEM(K=20)  & \textbf{94.2} & 8.72 & 12.2 \\
\midrule
  \multicolumn{4}{l}{Single Hybrid Model} \\
\midrule
Residual Flow~\cite{chen2019residual}         & 70.3 & 3.60  & 46.4 \\
IGEBM(K=60)~\cite{du2019implicit}      & 49.1 & 8.30  & 37.9 \\
JEM(K=20)$^+$~\cite{jem}    & 92.9 & 8.76  & 38.4 \\
JEM++(M=5)$^+$\cite{jempp}      & 91.1 & 7.81 & 37.9  \\
JEM++(M=10)     & 93.5 & 8.29 & 37.1  \\
JEM++(M=20)     & 94.1 & 8.11 & 38.0  \\
JEAT~\cite{jeat}             & 85.2 & \textbf{8.80} & 38.2 \\
\midrule
\multicolumn{4}{l}{EBM + Generator} \\
\midrule
 VERA($\alpha$=100) & 93.2 & 8.11 & 30.5 \\
 VERA($\alpha$=1)~\cite{nomcmc}   & 76.1 & 8.00 & 27.5 \\
\midrule
softmax              & 95.8  & -  & - \\
\bottomrule
\end{tabular}
\begin{tablenotes}
  \scriptsize\item * We report the best performance before the diverging of training. 
  \scriptsize\item $^+$ They suffer from high instability and regularly diverge.
\end{tablenotes}
\end{threeparttable}
\end{minipage}
\begin{minipage}[ht]{0.53\linewidth}

\begin{center}
\begin{threeparttable}\small
\caption{Hybrid Modeling Results on CIFAR100.}
\label{table:hybrid_cifar100}
\begin{tabular}{l|ccc}
\toprule
Model  &  Acc \% $\uparrow$ & IS $\uparrow$ & FID $\downarrow$  \\
\midrule
Softmax                & 78.9 & - & - \\
SNGAN(Cond)                                   & -    & 9.30 & 15.6  \\
BigGAN(Cond)                       & -    & 11.0 & 11.73 \\
\midrule
JEM(K=20)*           & 70.4 & 10.32 & 51.7 \\
JEM(K=30)*           & 72.8 & 10.84 & 34.2 \\
JEM++(K=5)*   & 72.0 & 8.19 & 37.7 \\
JEM++(K=10)*               & 74.5 & 10.23 & 32.9 \\
VERA($\alpha$=100)*  & 69.3 & 8.14  & 28.2  \\
VERA($\alpha$=1)*  & 48.7 & 7.97 & 26.6 \\
\midrule
M-JEM(K=1)$^+$  & 46.5  & 8.71 & 26.2 \\
M-JEM(K=2)$^+$  & 63.5 & 11.22 & 15.1 \\
M-JEM(K=5)  & 73.5 & \textbf{11.95} & 13.5 \\
M-JEM(K=10) & \textbf{75.1} & 11.72 & \textbf{12.7} \\
\bottomrule
\end{tabular}
\begin{tablenotes}
  \scriptsize\item * No official IS and FID scores are reported.
  \scriptsize\item $^+$ We report the best FID before diverging. 
\end{tablenotes}
\end{threeparttable}
\end{center}
\end{minipage}

As we can observe in Fig.~\ref{fig:tsne}, M-JEM with label information forms more compact manifolds of $\bs{x}^+$ and  $\bs{x}^-$ than M-EBM. In other words,  M-JEM-generated samples $\bs{x}^-$ match the distribution of real data and have lower variance and less \textbf{manifold intrusion}\cite{kmixup} than M-EBM. It gives us an explanation of why label information can improve generation quality.  Moreover, the latent feature space of M-JEM is better formulated than LSE-JEM, and there's less overlap between $\bs{x}^0$ and $\bs{x}^+$ which is desired since $\bs{x}^0$ and $\bs{x}^+$ should be assigned with different energies. 
Intuitively,  the number $K$ of SGLD sampling required for stable training is correlated to the distance between manifolds of $\bs{x}^0$ and $\bs{x}^+$. However, in Figures~\ref{fig:tsne} and ~\ref{fig:tsne_reg}, we can also observe that $\bs{x}^0$ and $\bs{x}^+$ are roughly mixed together from M-EBM without regularization and LSE-JEM.  Hence, it's interesting to reconsider the distance and the training instability when $\bs{x}^0$ and $\bs{x}^+$ are somewhat mixing together. We leave the exploration of this phenomenon as an existing direction for future.

\section{Conclusion}

In this paper, we propose simple yet effective training techniques to improve the image generation quality, training speed, and stability of unconditional EBM and JEM altogether. The experimental results demonstrate that our models achieve comparable performance on unconditional EBMs and JEMs, and enable us to scale the MCMC-based EBM learning to high-resolution large-scale image datasets, such as CelebA-HQ 128x128 and ImageNet 32x32 with the least MCMC sampling steps, making EBM training more practical for a research lab in academia to afford and explore.

%
%
\bibliographystyle{splncs04}
\bibliography{ml}

\appendix

\section{Experimental Details}\label{app:exp}

Our code is largely built on top of JEM~\cite{jem}\footnote{\url{https://github.com/wgrathwohl/JEM}}. 
For a fair comparison with JEM, we update each model with 390 iterations in 1 epoch. Empirically, we find a batch size of 128 for $p_{\bs{\theta}}(y|\bs{x})$ achieves the best classification accuracy on CIFAR10, while we use 64, the same batch size as in JEM, for the maximum likelihood estimate of $p_{\bs{\theta}}(\bs{x})$. All our experiments are performed with PyTorch on Nvidia GPUs. For CIFAR10 and CIFAR100, we train the backbone Wide-ResNet 28-10~\cite{wideresnet16} on a single GPU. Due to limited computational resources, we use Wide-ResNet 28-2 for ImageNet 32x32 on a single GPU, and Wide-ResNet 28-5 for CelebA-HQ 128x128 on 2 GPUs.

Table~\ref{table:hyperparameters} lists the hyper-parameters of our M-EBM/JEM algorithms. We train all our models for 200 epochs with the SGD optimizer, a buffer size of 10,000, a reinitialization frequency of 5\%. For CIFAR10, we use a larger learning rate of 0.1, while for CIFAR100, CelebA-HQ and Imagenet 32x32 we use an initial learning rate of 0.02. 

\begin{table}[ht!]
\caption{Hyper-parameters of M-EBM/JEM}
\label{table:hyperparameters}\vspace{-15pt}
\begin{center}
\begin{threeparttable}
\begin{tabular}{l|cc}
\toprule
Variable      & Value \\
\midrule
Epochs $N$                        & 200         \\
Buffer size $|\mathbb{B}|$       & 10,000      \\
Reinitialization freq. $\rho$    & 5\%         \\
SGLD step-size $\alpha$          & 1           \\
SGLD noise $\sigma$              & 0.001       \\
\bottomrule
\end{tabular}
\end{threeparttable}
\end{center}
\end{table}

\section{Training Speed}

We report the empirical training speeds of our M-JEM and baseline methods on a single Titan GPU in Table~\ref{table:speed}. As discussed previously, two mini-batches are utilized in M-JEM: one for training of EBMs and the other one for training of classifiers. In our experiments, we set the mini-batch size to 64 for EBM training, but use 128 for the classification batch because it achieves the best accuracy. It can be observed that M-EBM/JEM with the least number of MCMC sampling steps (e.g., $K=2$ and $K=5$)   can train without divergence\footnote{We can't guarantee the stability. M-EBM(K=2)/M-JEM(K=5) sometimes can train without failure, while other EBMs always diverge before 100 epochs}, and leads to 4.87x and 2.22x speedups over the original JEM, respectively. Here, the acceleration comes from the reduced MCMC sampling steps without stability dropping. M-JEM do not affect the sampling speed since it uses the same backbone and the SGLD sampling algorithm as JEM and its variants. Notably, Diffusion-based EBMs~\cite{diffusionRecovery} and VAEBM~\cite{vaebm}, requiring 8 GPUs/TPUs for days of training, are too expensive to be practical for a research lab in academia to afford.

\begin{table}[ht!]
\caption{Run-time comparison on CIFAR10. We set 390 iterations as one epoch and the training epochs are 200 epochs. The batch size is 64 for all models except DiffuRecov and VAEBM. }
\vspace{-10pt}

\label{table:speed}
\begin{center}

\begin{threeparttable}
\begin{tabular}{lccc}
\toprule
\multirow{2}{*}{Model}  &  Minutes    & Runtime  & Actual  \\
                   &  per Epochs & (Hours)  & Speedup \\
\midrule
Classifier         &  0.77  & 2.6   &                \\
JEM(K=20)$^{*}$    &  15.1  & 50.3  & $1 \times$     \\
JEM++(K=5)         &  6.3   & 20.9  & $2.39 \times$  \\
JEM++(K=10)        &  10.2  & 33.9  & $1.48 \times$  \\
VERA               &  9.6   & 32.2  & $ 2.8 \times$$^{\dagger}$   \\
\midrule
M-EBM   \\
K=1                &  2.4  & 7.9   & $6.29 \times$  \\
K=2                &  3.1  & 10.3  & $4.87 \times$  \\
K=5                &  5.4  & 18.0  & $2.79 \times$  \\
K=10               &  9.0  & 30.0  & $1.67 \times$  \\
\midrule
M-JEM   \\
K=1                &  3.8   & 12.7  & $3.97 \times$  \\
K=2                &  4.6   & 15.3  & $3.28 \times$  \\
K=5                &  6.8   & 22.7  & $2.22 \times$  \\
K=10               &  10.5  & 35.0  & $1.43 \times$  \\
\midrule
Other  \\
IGEBM              & \multicolumn{3}{c}{1 GPU for 2 days}              \\
KL-EBM             & \multicolumn{3}{c}{1 GPU for 1 days}        \\
DDPM               & \multicolumn{3}{c}{800k iter, 8 TPUs, 10.6 hours} \\
DiffuRecov         & \multicolumn{3}{c}{240k iterations, 8 TPUs, 40+ hours} \\
VAEBM$^+$          & \multicolumn{3}{c}{400 epochs, 8 GPUs, 55 hours} \\
\bottomrule
\end{tabular}
\begin{tablenotes}
 \scriptsize\item $^{*}$ JEM(K=20) is much less stable than M-EBM(K=2) and M-JEM(K=5).
 \scriptsize\item $^{\dagger}$ VERA reports a $2.8\times$ speedup while we run the official code and report a fair comparison results.
  \scriptsize\item $^+$ The runtime is for pretraining NVAE only. For VAEBM, they report the training takes around 25,000 iterations (or 16 epochs) on CIFAR-10 using one 32-GB V100 GPU. Then they cannot generate realistic samples anymore.
\end{tablenotes}
\end{threeparttable}
\vspace{-10pt}
\end{center}
\end{table}

\section{Qualitative Analysis of Samples}

Generation quality is difficult to qualify. Following the setting of JEM~\cite{jem}, we conduct a qualitative analysis of samples on CIFAR10.

\begin{figure*}[ht]

    \centering

    \subfigure[Samples with highest $p(x)$]{
        \includegraphics[width=0.22\columnwidth]{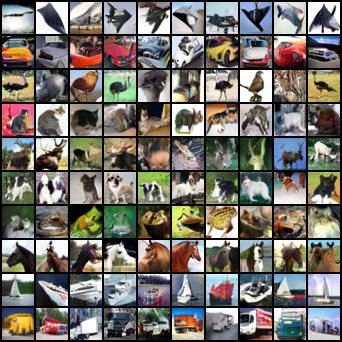} \label{figure:top_px}
    }
    \subfigure[Samples with lowest $p(x)$]{
        \includegraphics[width=0.22\columnwidth]{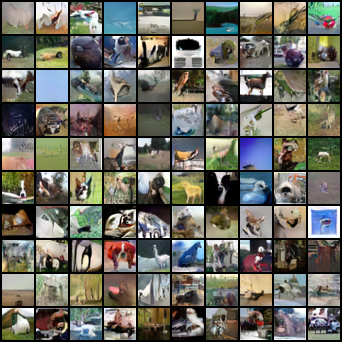} \label{figure:bottom_px}
    }
    \subfigure[Samples with highest $p(y|x)$]{
        \includegraphics[width=0.22\columnwidth]{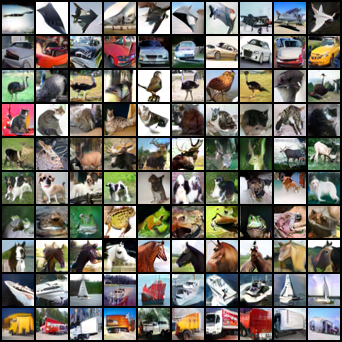} \label{figure:top_pyx}
    }
    \subfigure[Samples with lowest $p(y|x)$]{
        \includegraphics[width=0.22\columnwidth]{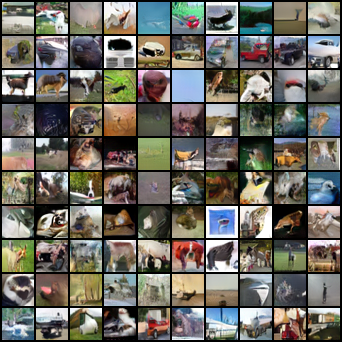} \label{figure:bottom_pyx}
    }
    \caption{Each row corresponds to 1 class.}
    \label{figure:categorial_topk_bottomk}

\end{figure*}

\begin{figure*}[ht]
    \centering
        \includegraphics[width=1\textwidth]{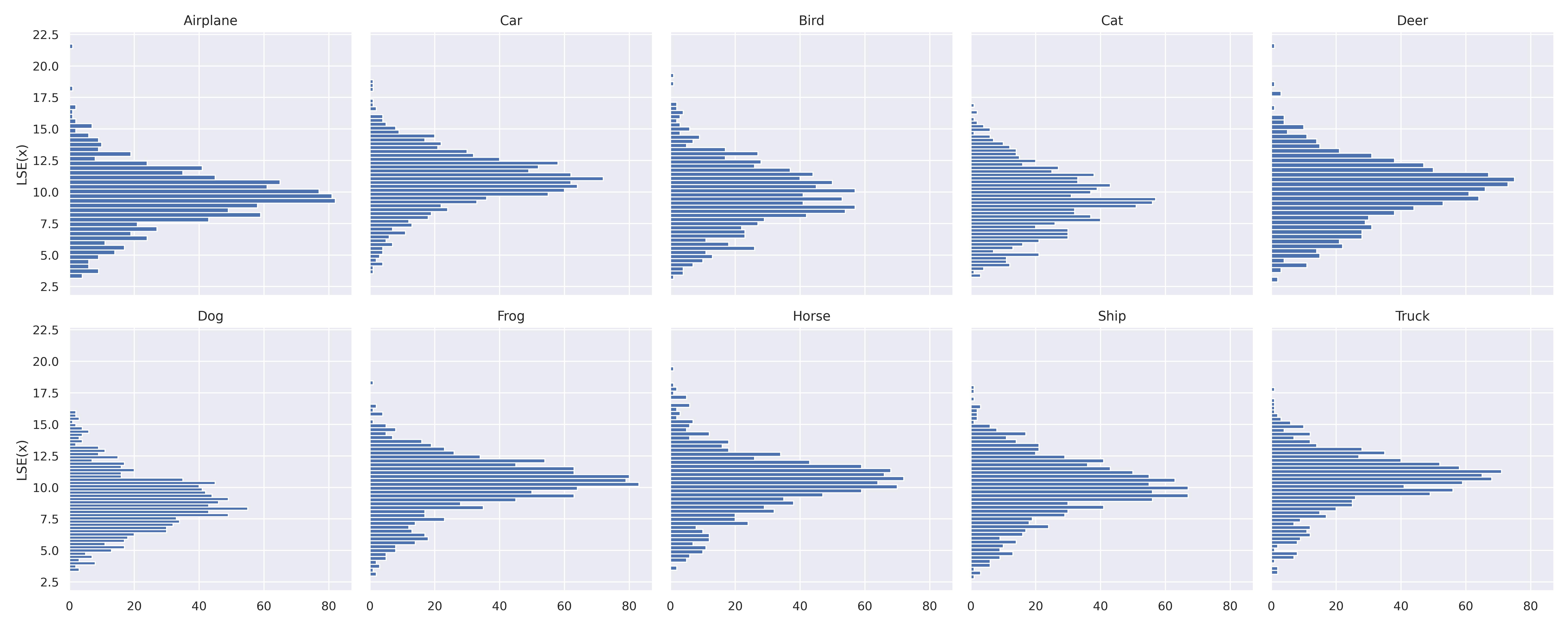}
    \caption{Histograms (oriented horizontally for easier visual alignment) of $\log p(x)$ arranged by class for CIFAR10.}
    \label{fig:px_category}
\end{figure*}

\section{Informative Initialization}
\begin{figure}[h]
    \centering
    \subfigure[CIFAR10]{
        \includegraphics[width=0.1\columnwidth]{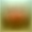}
        \label{figure:cifar10_center}
    }\hspace{10pt}
    \subfigure[CIFAR100]{
        \includegraphics[width=0.1\columnwidth]{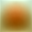}
        \label{figure:CIFAR100_center}
    }\hspace{10pt}
    \subfigure[Imagenet]{
        \includegraphics[width=0.1\columnwidth]{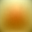}
        \label{figure:img32_center}
    }\hspace{10pt}
    \subfigure[CelebA-HQ]{
        \includegraphics[width=0.1\columnwidth]{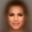}
        \label{figure:celeba128_center}
    }\vspace{-10pt}
    \caption{The centers ($\bs{\mu}$'s) of CIFAR10, CIFAR100, Imagenet 32 and CelebA-HQ 128.}
    \label{figure:informative_centers}
\end{figure}

In this work, we estimate the initial sampling distribution as
\begin{align}\label{eq:p0_app}
  p_0(\bs{x})&=\mathcal{N}(\bs{\mu}, \bs{\Sigma})\\
  \text{with}\quad 
\bs{\mu}  = \mathbb{E}_{\bs{x} \sim \mathcal{D} }[\bs{x}],  \quad 
\bs{\Sigma} &=\mathbb{E}_{\bs{x} \sim \mathcal{D} }\left[\left( \bs{x} - \bs{\mu} \right)\left( \bs{x} -\bs{\mu} \right)^{\top}\right],\nonumber
\end{align}
where $\mathcal{D}$ denotes the whole training set. Fig.~\ref{figure:informative_centers} illustrate the center ($\bs{\mu}$) estimated from each benchmark datatset.

\begin{figure*}[h]
    \centering
        \includegraphics[width=0.25\columnwidth]{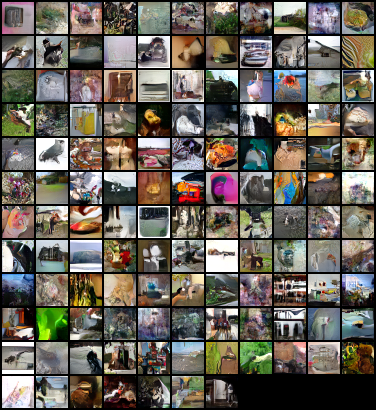}\vspace{-5pt}
    \caption{M-EBM generated samples of Imagenet 32x32.}
    \label{figure:celeba128_examples}
\end{figure*}

\section{Additional Generated Samples}

\begin{figure*}[ht!]
    \centering
    \subfigure[Samples with highest $p(\bs{x}$)]{
        \includegraphics[width=0.22\columnwidth]{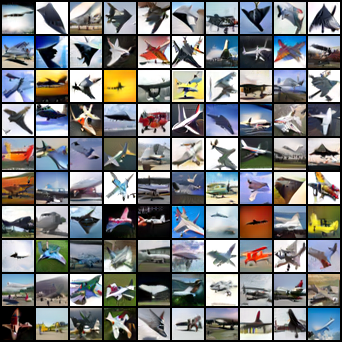}
    }
    \subfigure[Samples with lowest $p(\bs{x})$]{
        \includegraphics[width=0.22\columnwidth]{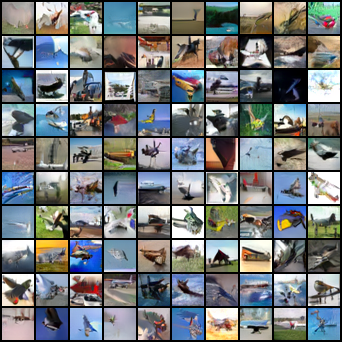}
    }
    \subfigure[Samples with highest $p(y|\bs{x}$)]{
        \includegraphics[width=0.22\columnwidth]{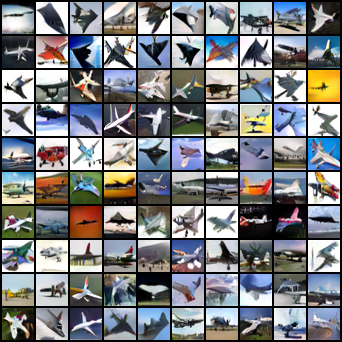}
    }
    \subfigure[Samples with lowest $p(y|\bs{x})$]{
        \includegraphics[width=0.22\columnwidth]{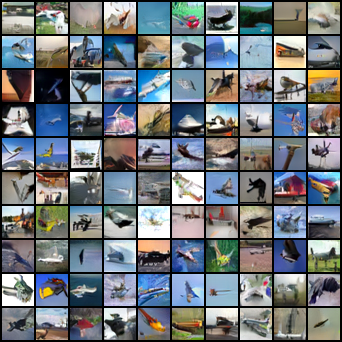}
    }
    \caption{M-JEM generated class-conditional samples of \textbf{Plane}}
    \label{figure:hybvit_app_class_0}
\end{figure*}

\begin{figure*}[ht!]
    \centering
    \subfigure[Samples with highest $p(\bs{x}$)]{
        \includegraphics[width=0.22\columnwidth]{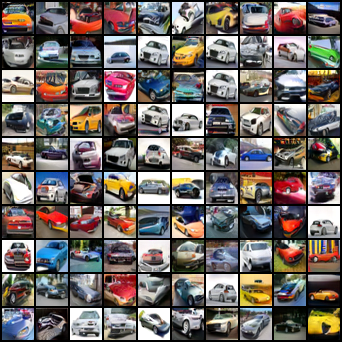}
    }
    \subfigure[Samples with lowest $p(\bs{x})$]{
        \includegraphics[width=0.22\columnwidth]{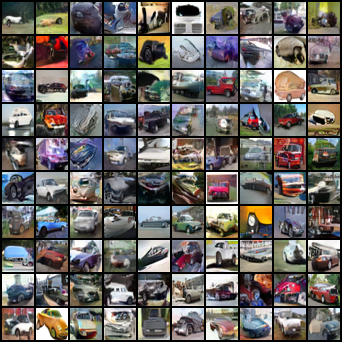}
    }
    \subfigure[Samples with highest $p(y|\bs{x}$)]{
        \includegraphics[width=0.22\columnwidth]{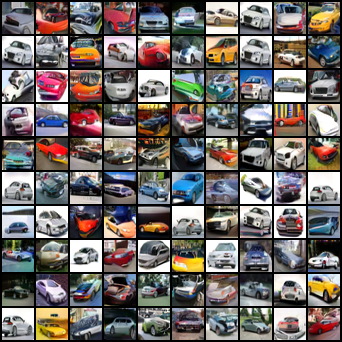}
    }
    \subfigure[Samples with lowest $p(y|\bs{x})$]{
        \includegraphics[width=0.22\columnwidth]{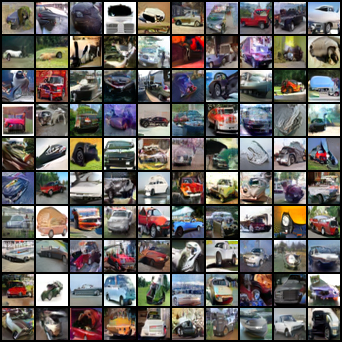}
    }
    \caption{M-JEM generated class-conditional samples of \textbf{Car}}
    \label{figure:hybvit_app_class_1}
\end{figure*}

\begin{figure*}[ht!]
    \centering
    \subfigure[Samples with highest $p(\bs{x}$)]{
        \includegraphics[width=0.22\columnwidth]{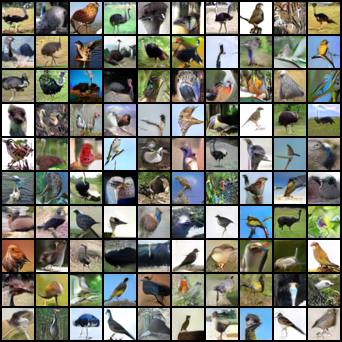}
    }
    \subfigure[Samples with lowest $p(\bs{x})$]{
        \includegraphics[width=0.22\columnwidth]{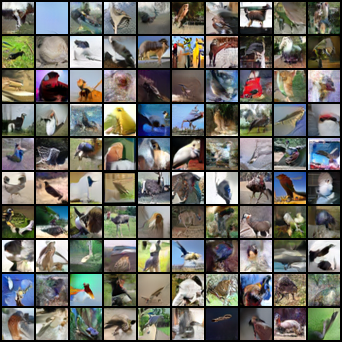}
    }
    \subfigure[Samples with highest $p(y|\bs{x}$)]{
        \includegraphics[width=0.22\columnwidth]{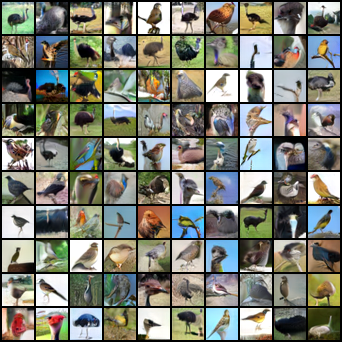}
    }
    \subfigure[Samples with lowest $p(y|\bs{x})$]{
        \includegraphics[width=0.22\columnwidth]{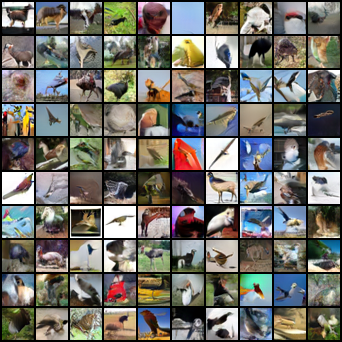}
    }
    \caption{M-JEM generated class-conditional samples of \textbf{Bird}}
    \label{figure:hybvit_app_class_2}
\end{figure*}

\begin{figure*}[ht!]
    \centering
    \subfigure[Samples with highest $p(\bs{x}$)]{
        \includegraphics[width=0.22\columnwidth]{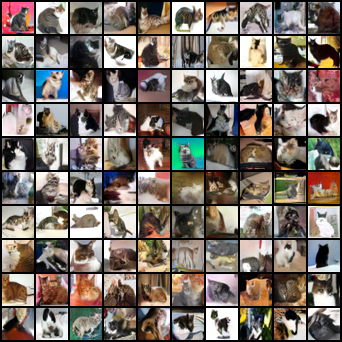}
    }
    \subfigure[Samples with lowest $p(\bs{x})$]{
        \includegraphics[width=0.22\columnwidth]{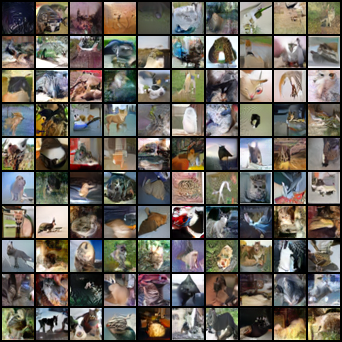}
    }
    \subfigure[Samples with highest $p(y|\bs{x}$)]{
        \includegraphics[width=0.22\columnwidth]{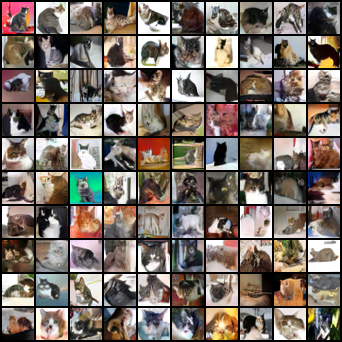}
    }
    \subfigure[Samples with lowest $p(y|\bs{x})$]{
        \includegraphics[width=0.22\columnwidth]{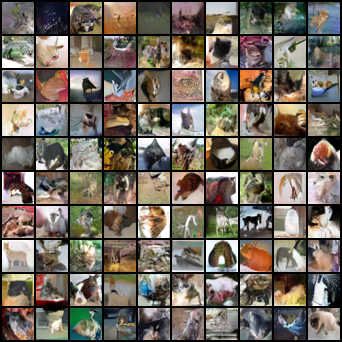}
    }
    \caption{M-JEM generated class-conditional samples of \textbf{Cat}}
    \label{figure:hybvit_app_class_3}
\end{figure*}

\begin{figure*}[ht!]
    \centering
    \subfigure[Samples with highest $p(\bs{x}$)]{
        \includegraphics[width=0.22\columnwidth]{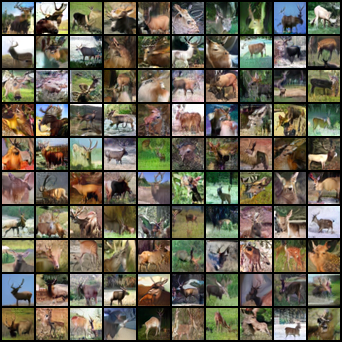}
    }
    \subfigure[Samples with lowest $p(\bs{x})$]{
        \includegraphics[width=0.22\columnwidth]{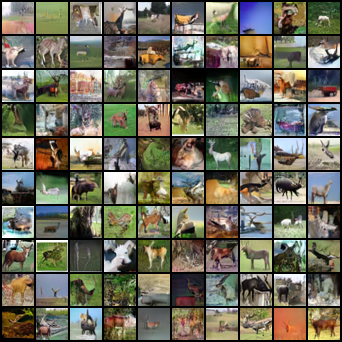}
    }
    \subfigure[Samples with highest $p(y|\bs{x}$)]{
        \includegraphics[width=0.22\columnwidth]{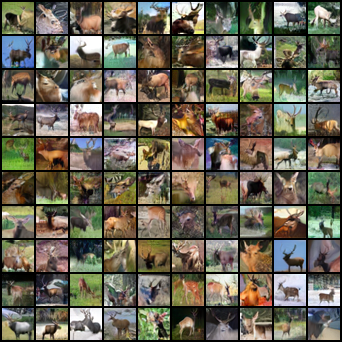}
    }
    \subfigure[Samples with lowest $p(y|\bs{x})$]{
        \includegraphics[width=0.22\columnwidth]{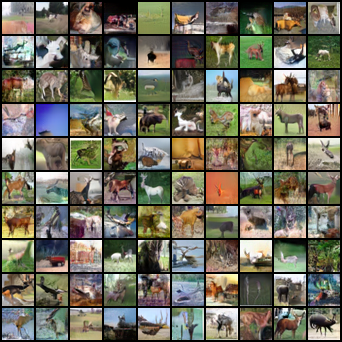}
    }
    \caption{M-JEM generated class-conditional samples of \textbf{Deer}}
    \label{figure:hybvit_app_class_4}
\end{figure*}

\begin{figure*}[ht!]
    \centering
    \subfigure[Samples with highest $p(\bs{x}$)]{
        \includegraphics[width=0.22\columnwidth]{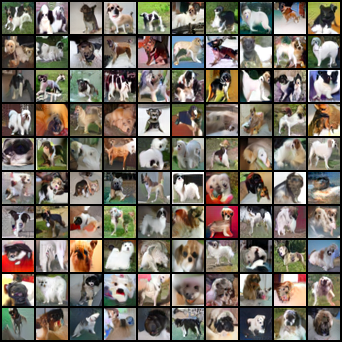}
    }
    \subfigure[Samples with lowest $p(\bs{x})$]{
        \includegraphics[width=0.22\columnwidth]{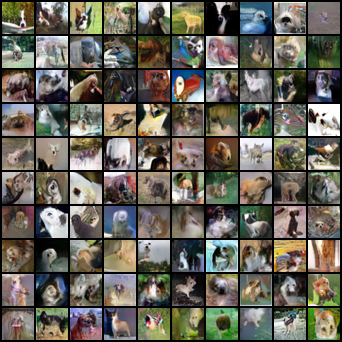}
    }
    \subfigure[Samples with highest $p(y|\bs{x}$)]{
        \includegraphics[width=0.22\columnwidth]{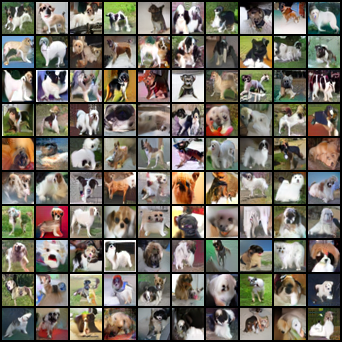}
    }
    \subfigure[Samples with lowest $p(y|\bs{x})$]{
        \includegraphics[width=0.22\columnwidth]{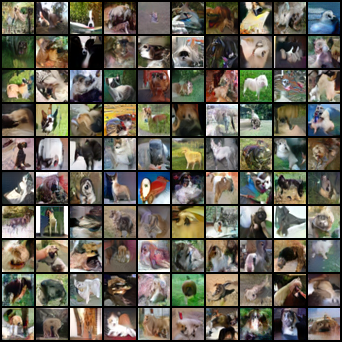}
    }
    \caption{M-JEM generated class-conditional samples of \textbf{Dog}}
    \label{figure:hybvit_app_class_5}
\end{figure*}

\begin{figure*}[ht!]
    \centering
    \subfigure[Samples with highest $p(\bs{x}$)]{
        \includegraphics[width=0.22\columnwidth]{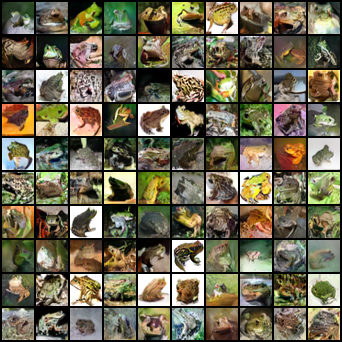}
    }
    \subfigure[Samples with lowest $p(\bs{x})$]{
        \includegraphics[width=0.22\columnwidth]{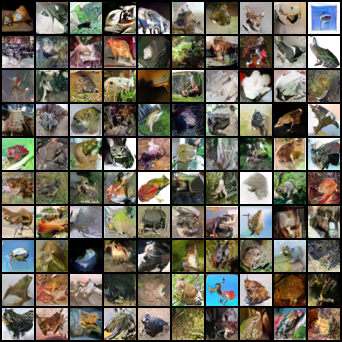}
    }
    \subfigure[Samples with highest $p(y|\bs{x}$)]{
        \includegraphics[width=0.22\columnwidth]{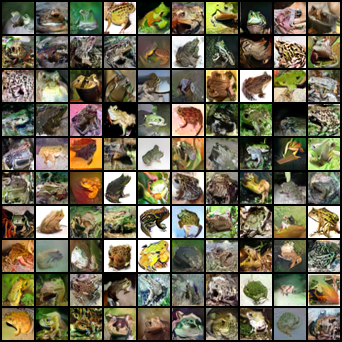}
    }
    \subfigure[Samples with lowest $p(y|\bs{x})$]{
        \includegraphics[width=0.22\columnwidth]{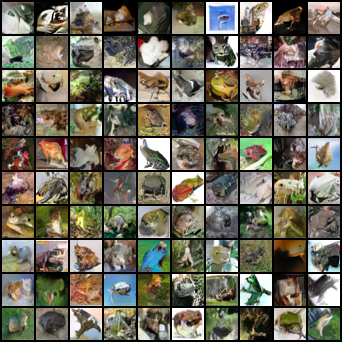}
    }
    \caption{M-JEM generated class-conditional samples of \textbf{Frog}}
    \label{figure:hybvit_app_class_6}
\end{figure*}

\begin{figure*}[ht!]
    \centering
    \subfigure[Samples with highest $p(\bs{x}$)]{
        \includegraphics[width=0.22\columnwidth]{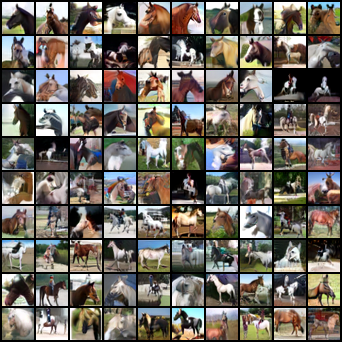}
    }
    \subfigure[Samples with lowest $p(\bs{x})$]{
        \includegraphics[width=0.22\columnwidth]{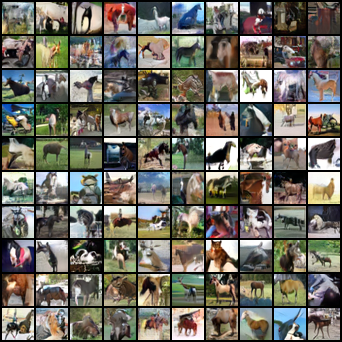}
    }
    \subfigure[Samples with highest $p(y|\bs{x}$)]{
        \includegraphics[width=0.22\columnwidth]{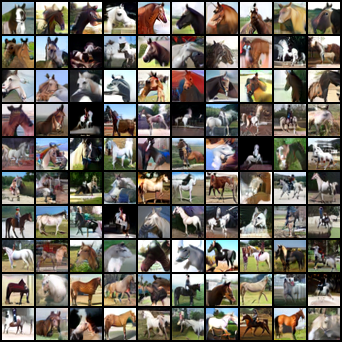}
    }
    \subfigure[Samples with lowest $p(y|\bs{x})$]{
        \includegraphics[width=0.22\columnwidth]{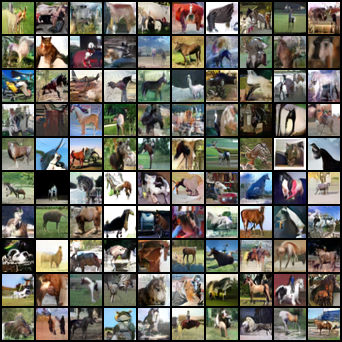}
    }
    \caption{M-JEM generated class-conditional samples of \textbf{Horse}}
    \label{figure:hybvit_app_class_7}
\end{figure*}

\begin{figure*}[ht!]
    \centering
    \subfigure[Samples with highest $p(\bs{x}$)]{
        \includegraphics[width=0.22\columnwidth]{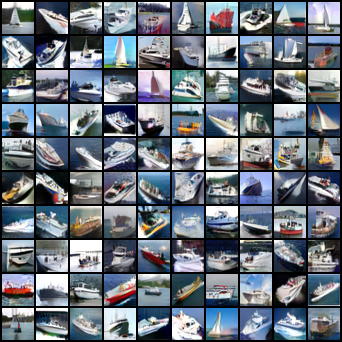}
    }
    \subfigure[Samples with lowest $p(\bs{x})$]{
        \includegraphics[width=0.22\columnwidth]{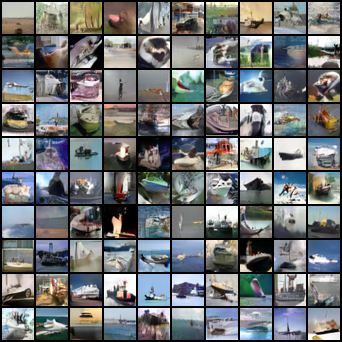}
    }
    \subfigure[Samples with highest $p(y|\bs{x}$)]{
        \includegraphics[width=0.22\columnwidth]{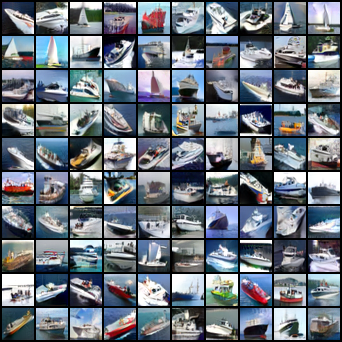}
    }
    \subfigure[Samples with lowest $p(y|\bs{x})$]{
        \includegraphics[width=0.22\columnwidth]{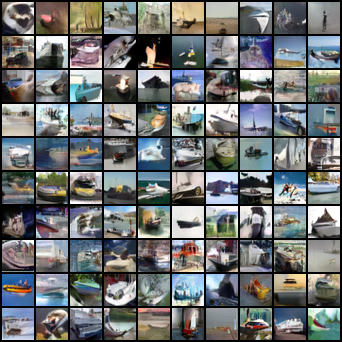}
    }
    \caption{M-JEM generated class-conditional samples of \textbf{Ship}}
    \label{figure:hybvit_app_class_8}
\end{figure*}

Fig.~\ref{figure:celeba128_examples} illustrates example images generated by M-EBM on Imagenet 32x32. Additional M-JEM generated class-conditional (best and worst) samples of CIFAR10 are provided in Figures~\ref{figure:hybvit_app_class_0}-~\ref{figure:hybvit_app_class_8}.

\end{document}